\documentclass[letterpaper, 10 pt, conference]{ieeeconf}
\IEEEoverridecommandlockouts
\usepackage{amsmath,amsfonts}
\usepackage{algorithmic}
\usepackage{algorithm}
\usepackage{array}
\usepackage[caption=false,font=normalsize,labelfont=sf,textfont=sf]{subfig}
\usepackage{textcomp}
\usepackage{stfloats}
\usepackage{url}
\usepackage{verbatim}
\usepackage{cite}
\usepackage{graphicx}
\graphicspath{{figures/}}

\usepackage{upgreek}
\usepackage{amssymb}
\usepackage[T1]{fontenc}
\usepackage{epstopdf}
\usepackage{amsmath, bm}
\usepackage[colorlinks,linkcolor=blue,anchorcolor=blue,citecolor=blue]{hyperref}
\usepackage{booktabs}
\usepackage{color,soul}
\usepackage{epsfig}
\usepackage{multirow}
\usepackage{mathptmx}
\usepackage{extarrows}
\usepackage{makecell}
\usepackage{ulem}
\usepackage{xcolor}
\usepackage{eso-pic}

\usepackage[authorsmultiline]{latexml}

\newcommand{\vect}[1]{\textbf{\textit{#1}}}

\newcommand{\matr}[1]{\textbf{#1}}

\newcommand{\fram}[1]{\left\{\mathrm{#1}\right\}}

\definecolor{ysyblue}{rgb}{0.1, 0.3, 0.8}

\title{\LARGE \bf
Lego-Like Stiffness Configuration of Planar Compliant \\ Modules for Task-Specific Flexible Interfaces}

\iflatexml
\author{%
\IEEEauthorblockN{Siyue Yao}%
\IEEEauthorblockA{State Key Laboratory of Mechanical Systems and Vibration, Shanghai Jiao Tong University, Shanghai 200240, China\\
Department of Mechanical Engineering, Stanford University, USA\\
\texttt{sy.yao@sjtu.edu.cn}}%
\and
\IEEEauthorblockN{Xiaochi Xie}%
\IEEEauthorblockA{Department of Mechanical Engineering, Stanford University, USA\\
\texttt{xxie15@stanford.edu}}%
\and
\IEEEauthorblockN{Shixuan Zhao}%
\IEEEauthorblockA{Department of Mechanical Engineering, Stanford University, USA\\
\texttt{sxuan@stanford.edu}}%
\and
\IEEEauthorblockN{Yutong Li}%
\IEEEauthorblockA{State Key Laboratory of Mechanical Systems and Vibration, Shanghai Jiao Tong University, Shanghai 200240, China\\
\texttt{Liyutong917@sjtu.edu.cn}}%
\and
\IEEEauthorblockN{Hao Li}%
\IEEEauthorblockA{Department of Mechanical Engineering, Stanford University, USA\\
\texttt{li2053@stanford.edu}}%
\and
\IEEEauthorblockN{Mark R. Cutkosky}%
\IEEEauthorblockA{Department of Mechanical Engineering, Stanford University, USA\\
\texttt{cutkosky@stanford.edu}}%
\and
\IEEEauthorblockN{Genliang Chen}%
\IEEEauthorblockA{Shanghai Key Laboratory of Intelligent Robotics; META Robotics Institute, Shanghai Jiao Tong University, Shanghai 200240, China\\
Corresponding: \texttt{leungchan@sjtu.edu.cn}}%
}

\else
\author{Siyue Yao$^{1,2}$, 
Xiaochi Xie$^{2}$, 
Shixuan Zhao$^{2}$, 
Yutong Li$^{1}$, 
Hao Li$^{2}$, \\
Mark R. Cutkosky$^{2}$, 
and Genliang Chen$^{3,4,*}$
\thanks{This work was supported in part by the National Natural Science Foundation of China under the Grant 525B2057 and the Science and Technology Commission of Shanghai Municipality under the Grant 25511104200.} 
\thanks{$^{1}$ 
State Key Laboratory of Mechanical Systems and Vibration, Shanghai Jiao Tong University, Shanghai 200240, China
({\tt\small sy.yao@sjtu.edu.cn}, {\tt\small Liyutong917@sjtu.edu.cn})}
\thanks{$^{2}$ Department of Mechanical Engineering, Stanford University, USA 
({\tt\small xxie15@stanford.edu}, {\tt\small sxuan@stanford.edu}, {\tt\small li2053@stanford.edu}, {\tt\small cutkosky@stanford.edu})}
\thanks{$^{3,4,*}$ 
Shanghai Key Laboratory of Intelligent Robotics;
META Robotics Institute, Shanghai Jiao Tong University, Shanghai 200240, China
(Corresponding: {\tt\small leungchan@sjtu.edu.cn}) }
}

\fi

\begin{document}
\AddToShipoutPictureBG*{\AtPageUpperLeft{\raisebox{-0.42in}{\makebox[\paperwidth][c]{\parbox{0.8\paperwidth}{\centering\footnotesize This work has been submitted to the IEEE for possible publication. Copyright may be transferred without notice, after which this version may no longer be accessible.}}}}}
\begin{sloppypar}

\maketitle
\thispagestyle{empty}
\pagestyle{empty}

\begin{abstract}
Compliant mechanisms provide compact and intrinsic structural compliance for regulating physical interactions between mechanisms and environments. 
However, different tasks demand distinct stiffness characteristics, often requiring task-specific optimization and redesign due to limited geometric design space and inherent coupling among multiple stiffness components. 
This paper presents a Lego-like stiffness configuration approach using stackable planar compliant modules. 
Three complementary module geometries are introduced, with their stiffness characteristics further regulated through beam width, plate thickness, and module orientation. 
A unified stiffness model is established for quantitative analysis of individual and composed modules. 
Further, a two-stage optimization method is presented to achieve desired stiffness profiles, combining a genetic algorithm for configuration and sequential quadratic programming for parameter refinement. 
Experimental verification shows deviations below $6.5\%$ for simulated stiffness. 
A flexible wrist is further developed as a representative implementation, exhibiting distinct compliant and dynamic responses under different stiffness characteristics. 
An optimized modular composition realizes prescribed stiffness values and maintains compliant obstacle interaction during high-speed motion at $1m/s$, with a maximum tested angular compliance of approximately $15^{\circ}$. 
The proposed framework provides a systematic approach for constructing flexible interfaces with task-specific stiffness characteristics. 
\end{abstract}
\iflatexml
\par\noindent\textbf{Funding.} This work was supported in part by the National Natural Science Foundation of China under the Grant 525B2057 and the Science and Technology Commission of Shanghai Municipality under the Grant 25511104200.
\par
\fi

\section{Introduction}\label{Sec::Intro}

\begin{figure}[!t]
	\centering
	\includegraphics[height=77mm]{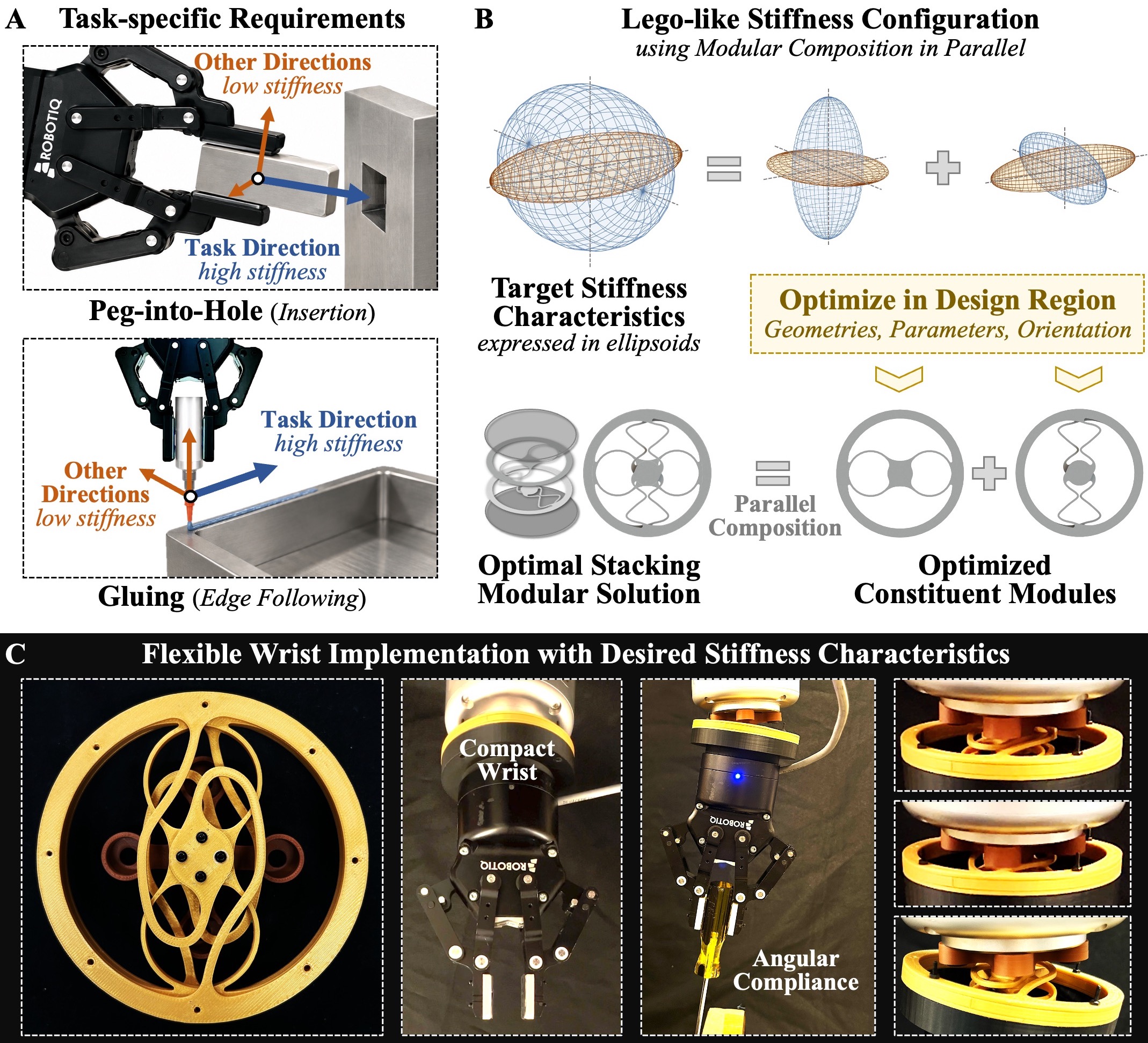}
	\caption{Overview of the proposed Lego-like stiffness configuration framework. 
    \textbf{(A)} Different tasks demand distinct stiffness requirements, especially differing along task and other directions. 
    \textbf{(B)} Stiffness configuration through optimization using parallel composition of planar compliant modules. 
    \textbf{(C)} Demonstrative flexible wrist implementation with desired stiffness. }
	\label{fig::sys_overall}
\end{figure}

Contact-rich manufacturing and industrial tasks, including assembly, surface finishing, and manipulation, require flexible interfaces to accommodate geometric uncertainties while maintaining desired contact conditions \cite{Whitney1982Assembly, Pagano2020Gluing}. %
However, different interaction configurations impose distinct stiffness requirements. 
As illustrated in Fig. \ref{fig::sys_overall}A, peg-in-hole and gluing impose different directional interaction requirements and consequently favor different stiffness configurations, with greater rigidity along motion or load transmission directions and increased compliance along accommodation directions\cite{Park2017PegInHole}. 
Moreover, the task-dependent stiffness is often demanded within a prescribed range, making task-specific and quantitatively defined stiffness configuration an important design requirement for flexible interfaces \cite{Mark1989Stiffness, Pollayil2023Impedance, Cutkosky1990}. 

Kinematics-based approaches are commonly employed for the synthesis of compliant mechanisms. 
The pseudo-rigid-body model (PRBM)\cite{Howell1994PRBM} facilitates design through equivalent rigid-link representations, while constraint-based design and Freedom and Constraint Topology (FACT) systematically construct flexure topologies according to desired degrees of freedom (DoFs) and constraints \cite{Hopkins2010FACT1, Awtar2007CBD, Yu2011ScrewTheory}. 
However, quantitative stiffness configuration requires further parametric optimization beyond the initial mechanism design.

Topology optimization provides another solution by directly optimizing material distribution within a prescribed design domain\cite{Sigmund1997Topology, Zhu2020TopologyReview}. 
Prescribed stiffness characteristics are incorporated into topology optimization by defining stiffness-based objectives and constraints. 
Zhang et al.\cite{Jin2016Topology, Zhu2018Decoupled, Liu2021ComplianceMatrix} developed a series of topology optimization formulations based on stiffness and Jacobian characteristics, enabling multi-DoFs synthesis, motion decoupling, and prescribed compliance while incorporating structural constraints. 
Further, Seltmann et al.\cite{Seltmann2023Distributed} incorporated stiffness and adaptive volume constraints to synthesize compliant mechanisms with distributed compliance and prescribed structural stiffness. 
However, the attainable design region is restricted by geometry constraints. 
Moreover, changes in stiffness requirements generally require redesign, which is often time-consuming. 

Beyond mechanism optimization as a whole, building-block approaches provide a higher-level design paradigm by composing predefined compliant elements\cite{Krishnan2012BuildingBlock}. 
Kim et al.\cite{Kim2008BuildingBlock} characterized compliant building blocks using stiffness and compliance ellipsoids and synthesized mechanisms through their combinations. 
DiBiasio et al. \cite{DiBiasio2012BuildingBlock} employed serial and parallel flexural blocks for rapid concept design. 
Li et al.\cite{Li2023CBE1} developed the Compliant Building Elements (CBE) method, in which a library of basic compliant elements is assembled in serial or parallel. 
These composition-based methods enable rapid mechanism design by assembling predefined and characterized elements, which can be designed through intuition, kinematics, or topology optimization\cite{Awtar2007FlexureModules, Blad2021BuildingBlock, McCarthy2023ConfigBlocks}. 
However, the design space remains constrained by the geometric restrictions within the mechanism.

To this end, this paper proposes a Lego-like stiffness configuration approach using stackable planar compliant modules, as illustrated in Fig. \ref{fig::sys_overall}B. 
By distributing stiffness contributions across multiple modules stacked along the thickness direction, the proposed framework introduces an additional design dimension that alleviates the geometric constraints of a single planar domain and enlarges the attainable stiffness range. 
With predefined module geometries, task-specific stiffness is configured through module selection, geometric parameters, orientation, and parallel composition. 
Three complementary planar modules, a unified stiffness model, and an optimization framework are developed and experimentally validated. 
A flexible wrist is further developed in Fig. \ref{fig::sys_overall}C as a representative implementation of flexible interfaces, showcasing the effectiveness of the proposed method in achieving desired stiffness characteristics. 

The main contributions include: 

(1) Proposed a Lego-like stacking strategy using planar compliant modules for task-specific stiffness regulation. 

(2) Established a unified stiffness model for parameterized modules and their parallel compositions, enabling quantitative characterization and analysis. 

(3) Developed a two-stage inverse design method for optimizing module configurations and parameters to achieve prescribed stiffness requirements. 

This paper is organized as follows. 
Planar compliant module designs, stiffness modeling, and optimization method are presented in Sec. \ref{Sec::Model}. 
Section \ref{Sec::Valid} experimentally characterizes the module stiffness to validate the simulated results. 
Implementation of a flexible wrist with an optimized modular composition for desired compliance and dynamic performance is shown in Sec. \ref{Sec::demos}. 
Section \ref{Sec::discuss} discusses the capabilities, limitations, and potential future extensions. 
Finally, conclusions are drawn in Sec. \ref{Sec::conclusion}. 

\section{Design and Modeling}\label{Sec::Model}
\subsection{Lego-Like Planar Compliant Module Design
}\label{SubSec::Design}

\begin{figure}[!t]
	\centering
	\includegraphics[height=120mm]{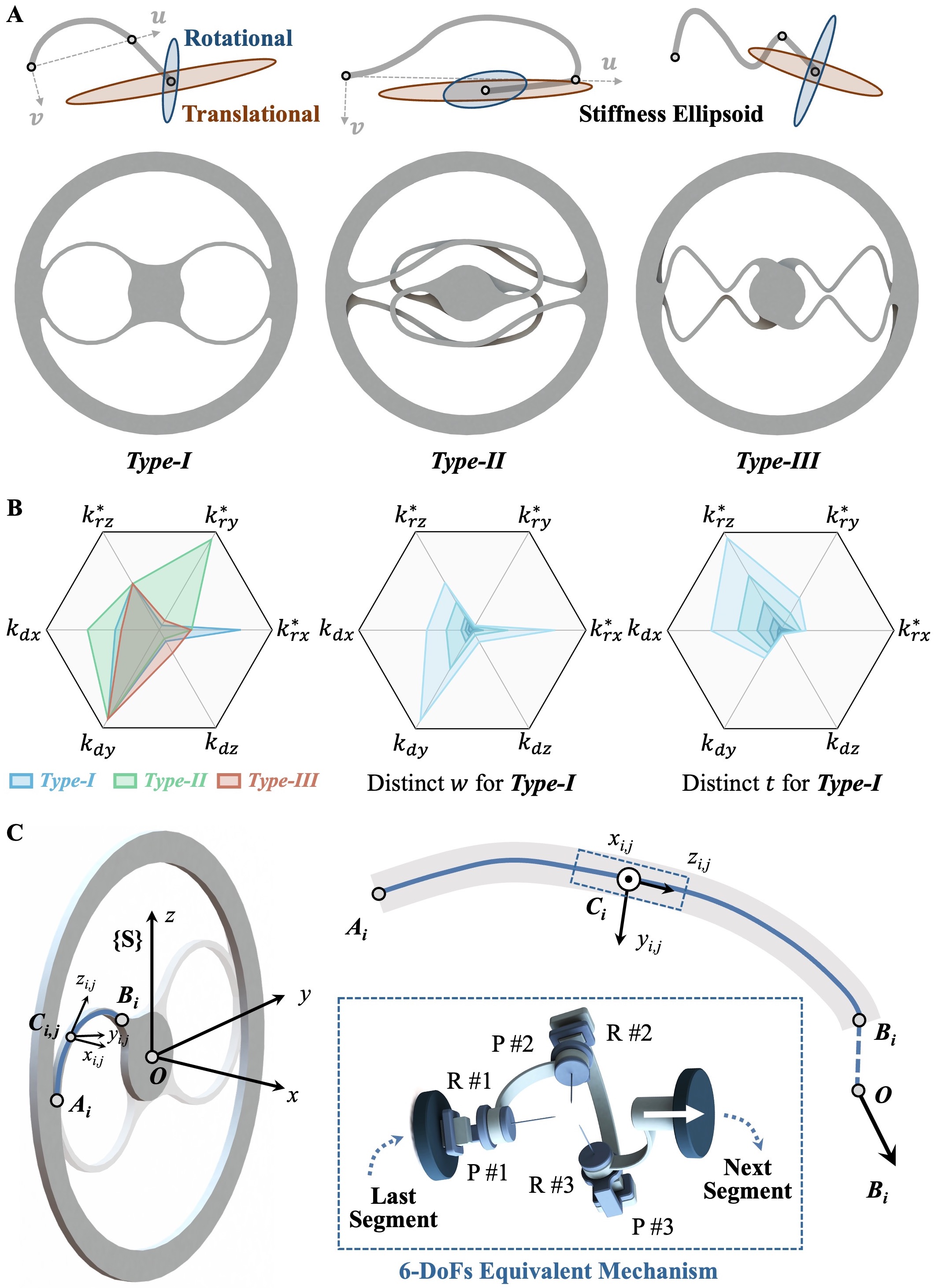}
	\caption{Planar compliant module design. 
    \textbf{(A)} Several types of modules with different geometries and complementary stiffness characteristics. 
    \textbf{(B)} Radar plot of normalized stiffness characteristics ($k^{*}_{ri} = 5 \cdot k_{ri} \cdot R^{-2}, i \in \{x,y,z\} $) of modules with different geometries, widths, and thicknesses. 
    \textbf{(C)} Discretization-based stiffness modeling method for one chain. }
	\label{fig::model}
\end{figure}

Decoupled stiffness characteristics between translation and rotation, and along different axes, are desired for manipulation and assembly interfaces, which reduce parasitic motion and provide a clear correspondence between task-level loading directions and the resulting structural deformation. 
Accordingly, each compliant module is constructed as a parallel mechanism with symmetrically distributed compliant chains, such that the off-diagonal terms of the stiffness matrix ideally vanish. 
One advantage of the proposed parallel composition is to allow a symmetric compliant mechanism with potential in-plane interference to be decomposed into two mirrored half-modules and stacked in separate planes, thereby preserving symmetry while expanding the design space. 
Further, the modules are designed as planar compliant mechanisms, with the plate thickness $t$ comparable to or larger than the in-plane chain width $w$. 
On this basis, the out-of-plane translational stiffness is treated as the load-support component rather than a principal configuration objective, which is required to remain within a task-dependent range. 

Three planar modules with distinct geometries are introduced, as illustrated in Fig. \ref{fig::model}A. 
Their geometries are intuitively devised based on the stiffness characteristics of curved beams. 
For a moderately curved beam, the in-plane translational stiffness along the axial direction $\vect{u}$, defined by the line connecting the two endpoints, is higher than that along the tangential direction $\vect{v}$, because deformation along $\vect{u}$ involves a greater contribution from axial stretching or compression, whereas deformation along $\vect{v}$ is dominated by bending.
In contrast, the in-plane rotational stiffness exhibits the opposite trend.
The difference between the two directional stiffnesses gradually diminishes as the beam curvature increases. 
Accordingly, the first module, \textit{Type-I}, is constructed using beams with relatively small curvature. 
As the curved beam extends into a spiral geometry, its stiffness characteristics, particularly the rotational components, change substantially.
Specifically, the rotational stiffness associated with the $\vect{u}$ direction becomes higher than that associated with $\vect{v}$. 
This observation motivates the second module, \textit{Type-II}, in which a curvature reversal is introduced. 
In addition, a higher-order wavy geometry is adopted for the compliant chains of \textit{Type-III}, which reduces the relative out-of-plane rotational stiffness compared with the in-plane ones. 
The illustrative stiffness characteristics of the three modules are depicted by the radar plots in Fig. \ref{fig::model}B using a common scale. 
These comparisons show that the three devised modules provide complementary combinations of the five principal stiffness components.

Besides the choices of module geometry, $w$ and $t$ serve as design parameters for tuning the stiffness characteristics. 
Since $w$ and $t$ influence the axial and bending stiffness at different rates, varying these parameters changes the relative ratios among the stiffness components. 
An illustration of tuning $w$ and $t$ for \textit{Type-I} is depicted in Fig. \ref{fig::model}B, where the different growth rates between each component are revealed, demonstrating the potential of achieving distinct stiffness characteristics through composition. 
The in-plane orientation of each module provides a further configuration variable, with a $90^\circ$ rotation used to swap the directional stiffness between the two in-plane axes. 
Accordingly, the three geometries, design variables $\{t, w\}$, and orientations jointly determine the stiffness properties. 
Task-specific stiffness profiles can then be regulated through module design and parallel composition, allowing the overall stiffness characteristics to be tailored toward the desired task requirements. 

The advantage of the proposed Lego-like planar compliant module composition strategy lies in separating topology design from stiffness configuration. 
With the module geometries predetermined, the time-consuming topology optimization is replaced by lower-dimensional parameter optimization for task-specific stiffness regulation. 

\subsection{Stiffness Modeling of Planar Compliant Modules
}\label{SubSec::Stiff_Model}

\begin{figure}[!t]
	\centering
	\includegraphics[height=85mm]{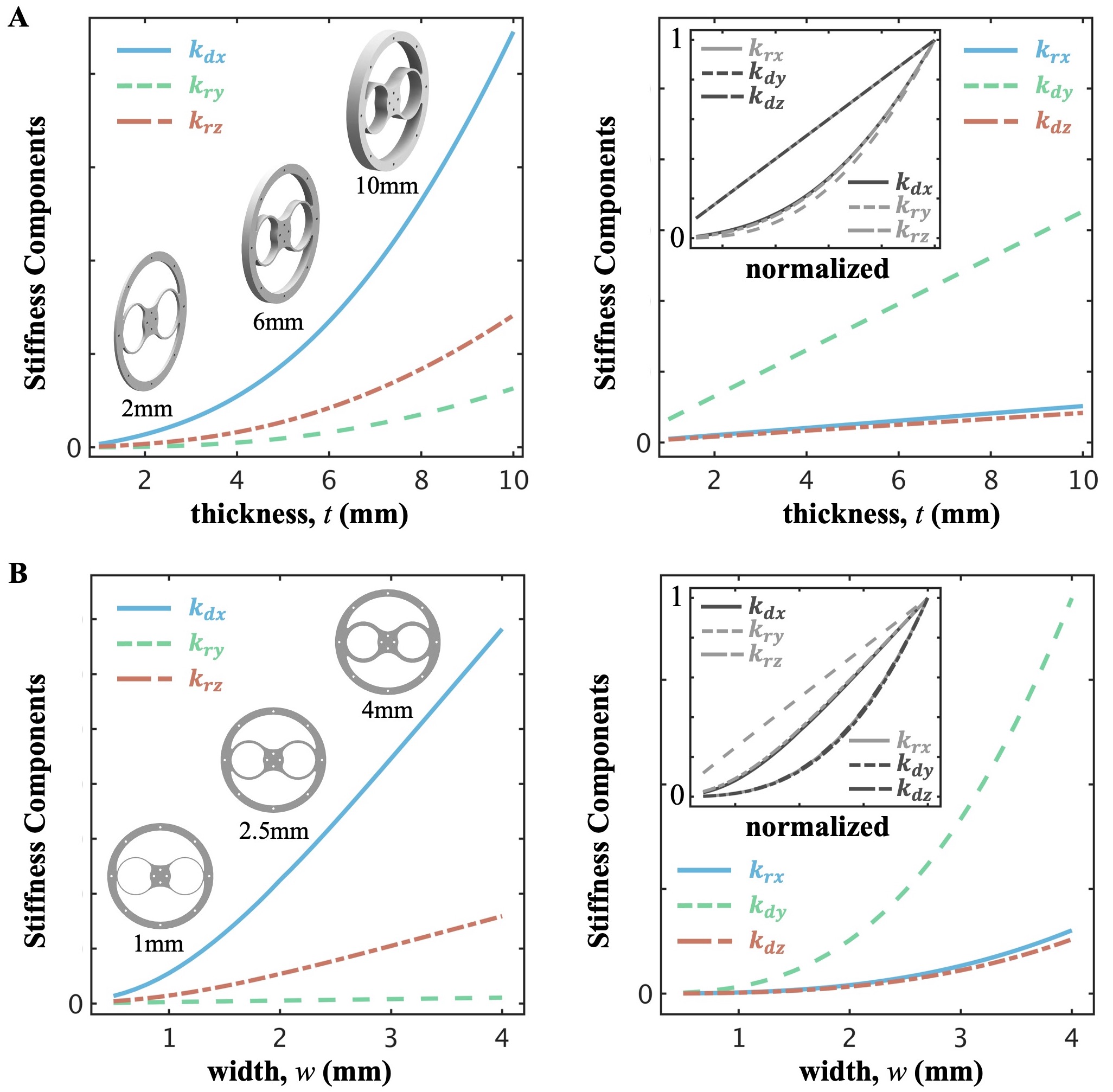}
	\caption{Variance of principal stiffness components with respect to the \textbf{(A)} plate thickness $t$ and \textbf{(B)} chain width $w$ for modules with \textit{type-I} geometry. }
	\label{fig::paras}
\end{figure}

To quantitatively characterize these parameterized modules and simulate their stiffness, a unified stiffness modeling method is applied based on discretization, equivalent mechanism representation, and compliance decomposition\cite{Chen2018Modeling}. 
As illustrated in Fig. \ref{fig::model}C, the ground reference frame $\fram{S}$ is established at the center point $O$ of the inner frame of the module.
The $x$-axis is defined along the normal direction of the module plane, while the $y$-axis is oriented vertically.
The $z$-axis is then determined according to the right-hand rule. 
Each curved compliant chain is discretized into $n$ segments with equal arc length. 
Each segment is further represented by a kinematically equivalent six-DoFs serial mechanism consisting of rigid links and elastic joints. 
Based on the local curvature, a local frame is established for the $j$-th segment of the $i$-th chain. 
Specifically, the $x_{i,j}$-axis remains parallel to the global $x$-axis, while the $z_{i,j}$-axis is aligned with the local tangential direction of the curve. 
The transformation from each local frame to $\fram{S}$ is incorporated when constructing the corresponding twists.
The kinetostatic equilibrium of the equivalent mechanism is subsequently established by combining geometric and force propagation constraints, resulting in a set of algebraic equations as

\vspace{-0.2cm}
\begin{equation}\label{equ::c}
    \vect{c} \left( \bm{\Theta}, \vect{F}_{t} \right) = \left[ \renewcommand\arraystretch{1.2}
    \begin{array}{c}
        \left( \mathrm{log} \left(
        \matr{g}_{\bm{\Theta}} \cdot
        \matr{g}_{t}^{-1} \right) \right)^{\vee} \\
        \vect{K}_{\bm{\Theta}} \cdot \bm{\Theta} - \matr{J}_{t, \bm{\Theta}}^{T} \cdot \vect{F}_{t} \\
    \end{array}\right] = \vect{0}
\end{equation}
where superscript Vee denotes the vector space representation of the log map of the Lie Group. 
$\bm{\Theta} \in \mathbb{R}^{6n \times 1}$ represents joint variables. 
$\matr{g}_{\bm{\Theta}} = \matr{T}_{\bm{\Theta}} \cdot \matr{g}_{0}$ where $\matr{T}_{\bm{\Theta}}$ is the transformation matrix. 
$\matr{g}_{0}$ and $\matr{g}_{t}$ represent the initial and target position and pose. 
$\matr{J}_{t, \bm{\Theta}}$ is the Jacobian matrix mapping from joint space to workspace. 
$\vect{F}_{t} \in \mathbb{R}^{6 \times 1}$ is the externally exerted forces and torques expressed in wrench form in $\fram{S}$. 
$\vect{K}_{\bm{\Theta}} \in \mathbb{R}^{6n \times 6n}$ denotes the structural stiffness matrix of elastic joints, whose torsional component utilizes the Saint-Venant constant. 

Eq. \eqref{equ::c} can be solved using the Newton-Laphson algorithm, with an analytical gradient. 
Furthermore, the stiffness matrix can be derived based on increment analysis\cite{Yao2026TMech}, which can be expressed as 

\vspace{-0.2cm}
\begin{equation}\label{equ::KS}
    ^{S}\vect{K} = ( \matr{J}_{t, \bm{\Theta}} \cdot 
    ( \vect{K}_{\bm{\Theta}} - \vect{K}_{\matr{J}} - 
    \vect{K}_{\matr{T}} )^{-1} \cdot \matr{J}_{t, \bm{\Theta}}^{T} )^{-1}
\end{equation}
where $\vect{K}_{\matr{J}}$ and $\vect{K}_{\matr{T}}$ represent the influence of the configuration and the relative position on the transformation of force. 

Eventually, the overall stiffness matrix is composed by adding up the stiffness matrix of each chain that are expressed in the same frame. 
It should be noted that Eq. \eqref{equ::KS} provides an analytic expression of the spatial stiffness matrix under a general end-loaded case. 
For initial balanced mechanism with no loads, $\vect{K}_{\matr{J}} = \vect{K}_{\matr{T}} = \matr{0}$. 
The expression can then be simplified as 
$^{S}\vect{K} = ( \matr{J}_{t, \bm{\Theta}} \cdot \vect{K}_{\bm{\Theta}}^{-1} \cdot \matr{J}_{t, \bm{\Theta}}^{T} )^{-1}$. 

Since the stiffness matrix of the module is diagonal due to the symmetric layout, the six principal stiffness components are extracted to represent the overall characteristics. 
It should be noted that the rotational stiffness is uniformly expressed as $k_{ri} \cdot R^{-2}$ ($i \in \{x,y,z\}$), where $R$ denotes the characteristic length, taken as the radius of the outer frame. 
The influence of $w$ and $t$ is simulated using \textit{Type-I} as the module geometry. 

Figure \ref{fig::paras}A shows the stiffness variance when $w = 1 \, mm$ while $t$ increases from $1 \, mm$ to $10 \, mm$. 
The six components can be divided into two groups. 
The in-plane components $k_{r_x}$, $k_{d_y}$, and $k_{d_z}$ increase approximately linearly with $t$, whereas $k_{d_x}$, $k_{r_y}$, and $k_{r_z}$ exhibit an approximately cubic dependence. 
This difference can be revealed from the axial and bending stiffness of the equivalent segment. 
Each stiffness component is further normalized by its corresponding maximum value among the simulated cases. 
The normalized components within each group exhibit nearly overlapping trends, which will be more evident when $w \ll t$, as the deformation more closely follows an ideal slender rectangular beam. 
A complementary trend is observed when varying $w$ from $0.5 \, mm$ to $4 \, mm$ with $t = 2 \, mm$, as shown in Fig. \ref{fig::paras}B. 
In this case, $k_{r_x}$, $k_{d_y}$, and $k_{d_z}$ show a pronounced nonlinear increase close to a cubic relationship, while the remaining components exhibit a linear to quadratic dependence on $w$. 
Similar overlap is observed after normalization. 
This deviation arises because $w$ and $t$ are about the same level. 
The linear relationship is maintained for $k_{d_x}$, $k_{r_y}$, and $k_{r_z}$ when $t \ll w$ holds. 

These simulation results demonstrate that tuning $w$ and $t$ substantially regulates both the magnitudes and relative ratios of the stiffness components. 
However, components governed by similar deformation modes exhibit coupled variations, such as the approximately linear and cubic components. 
Therefore, the six components cannot be independently tuned using a single module geometry, which leads to the composition of \textit{Type-II} and \textit{Type-III}. 
Each type of module possesses unique stiffness characteristics, such that a broader range of stiffness profiles can be obtained through parallel composition. 
Moreover, the relationship between the module parameters and the resulting stiffness is nonlinear, making task-specific stiffness regulation rely on joint optimization of the module parameters, geometry, and orientation. 

\subsection{Optimization for Stiffness Configuration} \label{SubSec::Opt}

\begin{figure}[!t]
	\centering
	\includegraphics[height=80mm]{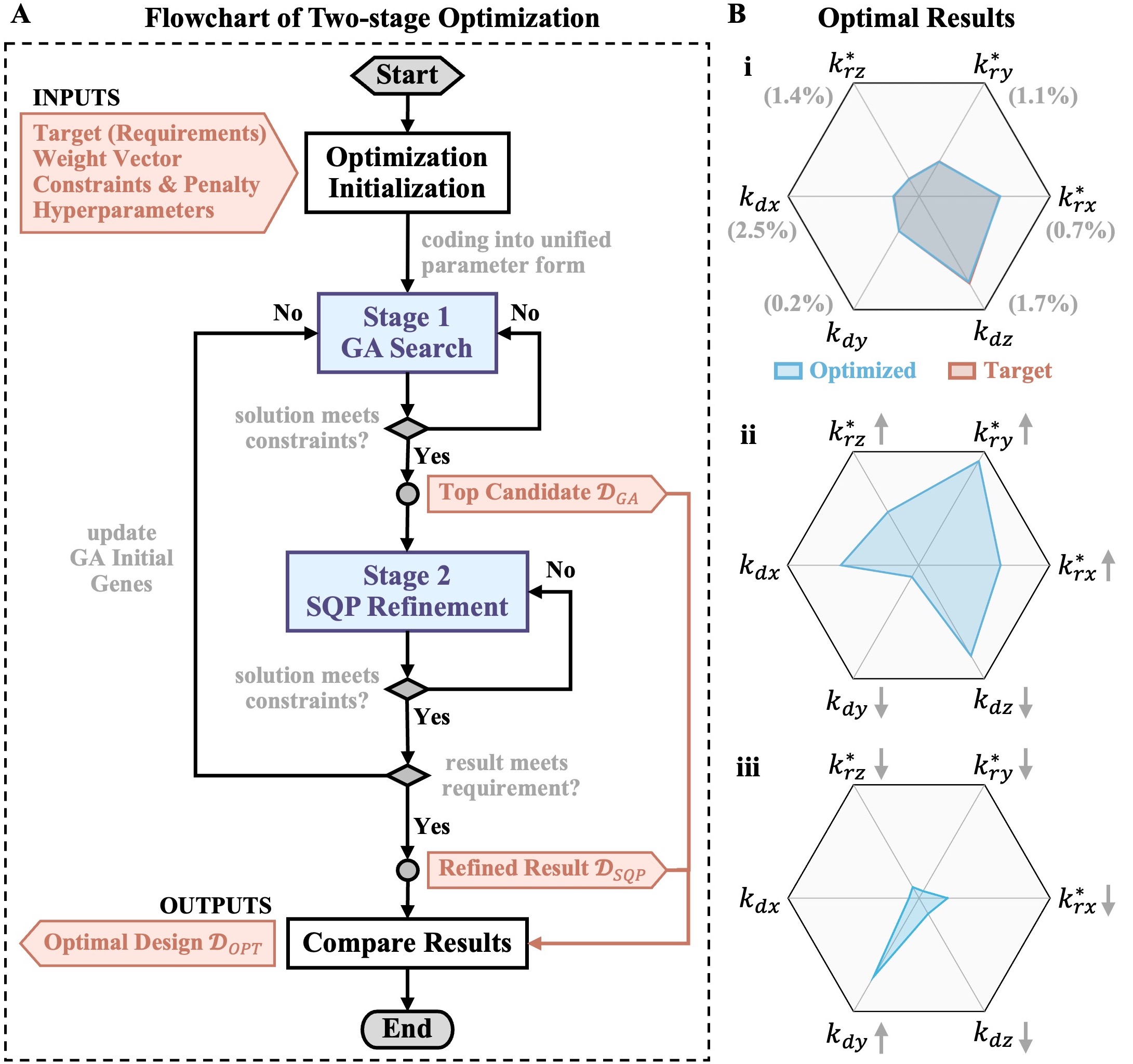}
	\caption{Optimization of module combination for stiffness configuration. 
    \textbf{(A)} Flowchart of the two-stage strategy for solving the optimization design problem. 
    \textbf{(B)} Stiffness radar plots of optimal results under different requirements using a same scale. %
    \textbf{(i)} Given target stiffness components. 
    \textbf{(ii)} Requiring higher rotational stiffness and lower translational stiffness. 
    \textbf{(iii)} Requiring a higher $k_{dy}$ and lower the remaining components other than $k_{dx}$. }
	\label{fig::opt}
\end{figure}

The stiffness configuration is formulated as an inverse design problem, in which the module types, dimensions, orientations, and number of stacked modules are considered within the same optimization framework. 
Given a target stiffness matrix with diagonal vector $\vect{K}_{t} = [k_{rx}, k_{ry}, k_{rz}, k_{dx}, k_{dy}, k_{dz} ]^{T}$, the objective is to determine the optimal module composition minimizing the stiffness deviation while satisfying given geometric and stiffness constraints. 
The complete optimization problem is expressed as 

\vspace{-0.2cm}
\begin{equation}\label{equ::OPT}
    \renewcommand{\arraystretch}{1.2}
    \begin{array}{rll} 
        \displaystyle \min_{\mathcal{D}} \quad &
        \multicolumn{2}{l}{ c(\mathcal{D}) = c_{0} \left( \vect{K}(\mathcal{D}), \vect{K}_{t} \right) + 
        \lambda \cdot \bm{\Phi} (\mathcal{D}) } \\
        \mathrm{s.t.} \quad 
        & t_{i} = t_{\min} + \tilde{t}_{i} \cdot 
        (t_{\max} - t_{\min}), 
        & \tilde{t}_{i} \in [0,1] \\
        & w_{i} = w_{\min} + \tilde{w}_{i} \cdot 
        (w_{\max} - w_{\min}), 
        & \tilde{w}_{i} \in [0,1] \\
        & a_{i} \in \{0, 1\}, \\
        & m_i\in\{1,2,3\}, \\ 
        & rot_{i} \in \{0, 1\}, \\ 
        & \multicolumn{2}{l}{ \vect{K}(\mathcal{D}) = \sum_{i=1}^{N} a_{i} \cdot \mathcal{F} \left( m_{i}, \tilde{t}_{i}, \tilde{w}_{i}, rot_{i} \right). } \\
    \end{array}
\end{equation}
where $i = 1, 2, ..., N$ denotes the index for each compliant module and $N$ is the total number. 
$N_{\max}$ denotes the maximum number of modules for composition. 
$\mathcal{D} = \{ \mathcal{D}_{1}, \mathcal{D}_{2}, ..., \mathcal{D}_{N_{\max}} \}$ represents one possible stacking solution, where $\mathcal{D}_{i}$ includes a set of parameters as $\{ a_{i}, m_{i}, \tilde{t}_{i}, \tilde{w}_{i}, rot_{i} \}$. 
$a_{i}$ denotes whether the module is active or not. 
$m_{i}$ stands for the three types with different geometries. 
$\tilde{t}_{i}$ and $\tilde{w}_{i}$ are the normalized thickness and width. 
Both parameters are confined to a reasonable range that is easy to manufacture. 
$rot_{i}$ represents the orientation of the module. 
The stiffness model is encapsulated into $\mathcal{F} \left( m_{i}, \tilde{t}_{i}, \tilde{w}_{i}, rot_{i} \right)$, which provides the principal characteristics under unloaded case. 
It should be further noted that the rotational stiffness components $k_{ry}$ and $k_{rz}$ are affected by the stacking thickness and sequence. 
Since the module is offset along the $x$-direction from the rotation center after stacking, its translational stiffness contributes additional rotational contributions. 
These contributions are relatively small for thin stacks but increase quadratically with the offset as the stack thickness increases. 

Furthermore, the objective function $c(\mathcal{D})$ consists of a task-dependent objective term and a penalty term. 
The form of $c_{0} \left( \vect{K}(\mathcal{D}), \vect{K}_{t} \right)$ depends on the stiffness requirement. 
For a prescribed target stiffness vector $\vect{K}_{t}$, it can be defined as $c_{0} = \left\| \vect{q}^{T} \cdot \left( \vect{K}(\mathcal{D}) - \vect{K}_{t} \right) \right\|_2$. 
For qualitative or relative stiffness requirements, it can instead be expressed as $c_{0} = \mathcal{G} \left( \vect{q}, \vect{K}(\mathcal{D}) \right)$. 
In both cases, $\vect{q}$ is a weight vector. 
Constraint violations are incorporated through hinge-loss penalties as $\bm{\Phi}_{(*)} = \max \{0, s \cdot \left( (*) - (*)_{th} \right) \}$, where $(*)$ denotes the constrained quantity and $s \in \{-1,1\}$ is selected according to whether an upper or lower bound is imposed. 
The total penalty $\bm{\Phi}(\mathcal{D}) = \bm{\Phi}_{N} + \bm{\Phi}_{T} + \bm{\Phi}_{K}$ consists of geometric and stiffness terms. 
$\bm{\Phi}_{N}$ limits the number of modules, 
$\bm{\Phi}_{T}$ constrains the total stack thickness, 
and $\bm{\Phi}_{K}$ enforces stiffness bounds, such as a minimum load-support stiffness. 
The coefficient $\lambda$ determines the penalty weight. 

Eq. \eqref{equ::OPT} is a Mixed-Integer Nonlinear Programming problem (MINLP), where $N$, $a_{i}$, $m_{i}$, and $rot_{i}$ must be taken as integers. 
The flowchart of the solving process is illustrated in Fig. \ref{fig::opt}A. 
A two-stage optimization strategy is employed, where a genetic algorithm (GA) determines integers for solution candidates, followed by sequential quadratic programming (SQP) to refine only the continuous parameters $\tilde{t}_{i}$ and $\tilde{w}_{i}$. 

Several optimal results under different prescribed requirements are demonstrated in Fig. \ref{fig::opt}B. 
The common optimization constraints are set as $N_{\max} = 6$ and $t_{\max} = 30 \, mm$, while the geometric parameters of each module are bounded by $t_{i} \in [1, 10] \, mm$ and $w_{i} \in [1, 3] \, mm$. 
A specific target stiffness is first prescribed as 
$\vect{K}_{t} = [40, 20, 10, 25000, 40000, 100000]^{T}$ expressed in $N \cdot m/rad$ and $N/m$.
The optimal combination consists of three modules: 
two $\textit{Type-I}$ with $\{t_{1}, w_{1}\} = \{ 4.9 , 1.6\} \, mm$ and $\{t_{2}, w_{2}\} = \{ 1.3 , 1.9\} \, mm$, both oriented at $90^{\circ}$, along with a $\textit{Type-III}$ with $\{t_{3}, w_{3}\} = \{ 1 , 2.8\} \, mm$ at $0^{\circ}$. 
The radar plot of the stiffness components is shown in Fig. \ref{fig::opt}B(i), where the rotational ones are normalized for a clearer comparison. 
The errors between the optimal result and the target are $[0.7, 1.1, 1.4, 2.5, 0.2, 1.7] \, \%$, which showcases the effectiveness of the proposed method. 

Furthermore, qualitative requirements are considered, where selected stiffness components are specified to be minimized or maximized rather than assigned specific values. 
Figure \ref{fig::opt}B(ii) demonstrates the results for a requirement of higher rotational stiffness and lower translational stiffness. 
The optimal combination is one $\textit{Type-II}$ with $\{t_{1}, w_{1}\} = \{ 5, 1\} \, mm$ and one $\textit{Type-III}$ with $\{t_{2}, w_{2}\} = \{ 10, 1.5\} \, mm$, both oriented at $90^{\circ}$. 
Moreover, a requirement of a higher $k_{dy}$ and lower values of the remaining stiffness components yields a three-module solution comprising one \textit{Type-II} module with $\{t_{1}, w_{1}\} = \{ 5.6, 2.8\} \, mm$ and one \textit{Type-III} module with $\{t_{3}, w_{3}\} = \{ 1.8, 2.9\} \, mm$, both oriented at $0^{\circ}$. 
The corresponding radar plot is shown in Figure \ref{fig::opt}B(iii). 
The results show that higher rotational stiffness with lower translational stiffness is favored by thicker yet narrower modules, whereas the opposite case is better achieved through the parallel composition of wider and thinner modules. 

\begin{figure}[!t]
	\centering
	\includegraphics[height=75mm]{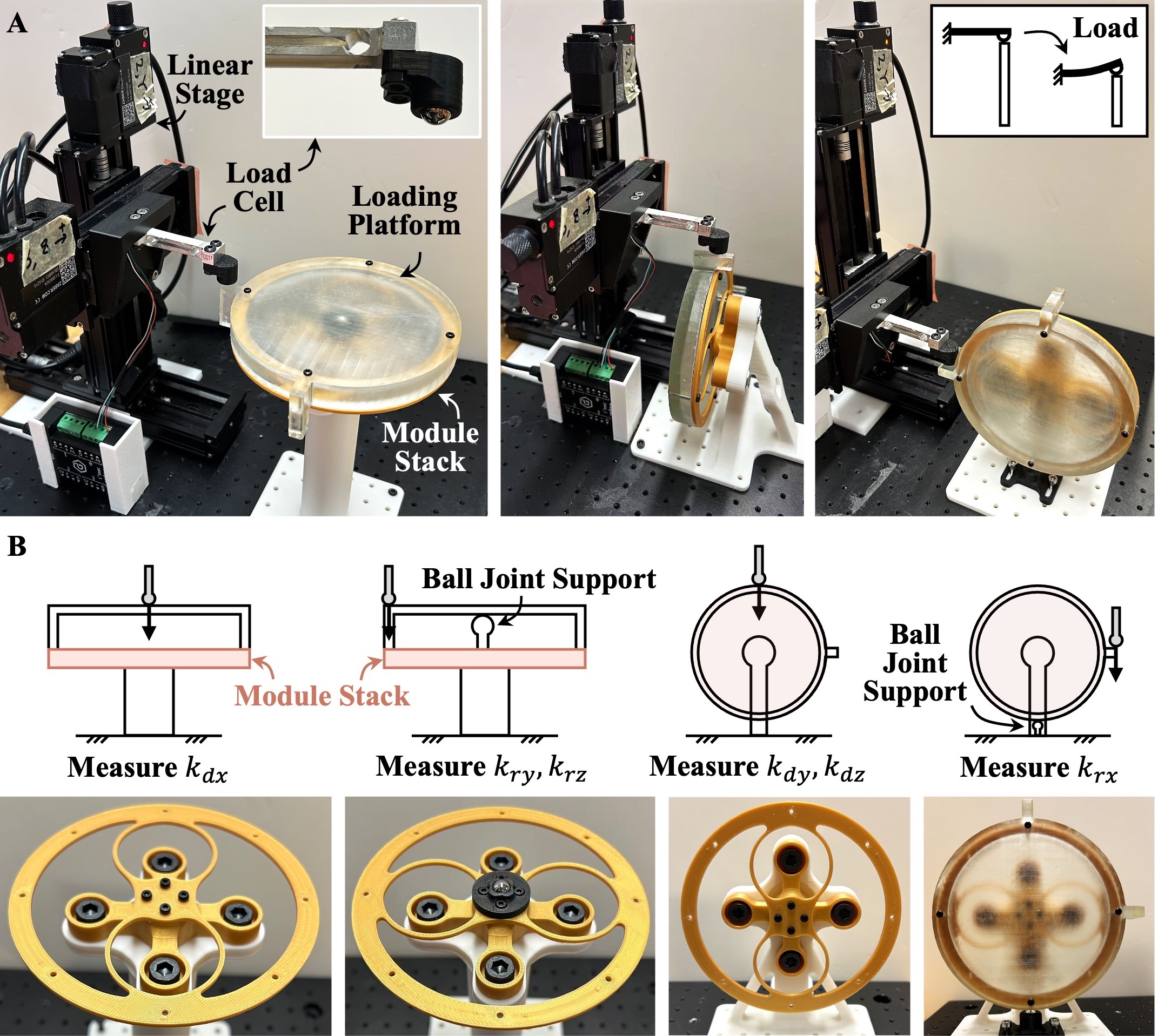}
	\caption{Stiffness validation using a \textit{Type-I} module. 
    \textbf{(A)} Experimental setup. 
    \textbf{(B)} Measurement configurations for the six principal stiffness components. }
	\label{fig::stiffExp}
\end{figure}

The resulting profile demonstrates that the optimization effectively achieves the prescribed qualitative stiffness preference. 
It should be noted that all optimal solutions are obtained on a hundred-second-level time scale, enabling time-efficient optimization for rapid design iterations. 

\section{Validation and Evaluation}\label{Sec::Valid}

\subsection{Experiment Setup}\label{SubSec::setup}

Experiments are conducted to validate the established stiffness model. 
A \textit{Type-I} compliant module with $\{t, w\} = \{ 2 , 1.5\} \, mm$ and $ R = 50 \, mm$ is fabricated and tested. 
The experimental setup is shown in Fig. \ref{fig::stiffExp}A. 
The six principal stiffness components are measured separately using the configurations illustrated in Fig. \ref{fig::stiffExp}B. 
Two fixture arrangements are employed to accommodate the different tested stiffness components. 
A motorized linear stage applies prescribed displacements to the loading platform, while recording the displacement at $30 \, Hz$. 
Meanwhile, the corresponding reaction force is recorded by a one-dimensional load cell with high precision at $200 \, Hz$. 
A ball joint is installed at the loading interface to allow self-alignment of the force direction and to reduce friction and parasitic constraints. 
For translational stiffness measurements, the load is applied through the center of the module such that the imposed displacement primarily produces translation along the target direction. 
For rotational stiffness measurements, the load is applied at an eccentric position with a determined lever arm relative to the center. 
Consequently, the measured displacement and force can be converted into the corresponding angular displacement and torque. 
To suppress parasitic translation during rotational loading, an additional ball-joint support is placed at the module center, constraining the translational motion while allowing the intended rotation. 
By changing the module orientation and fixture configuration, the same setup is used to characterize the stiffness components. 

\begin{figure}[!t]
	\centering
	\includegraphics[height=66mm]{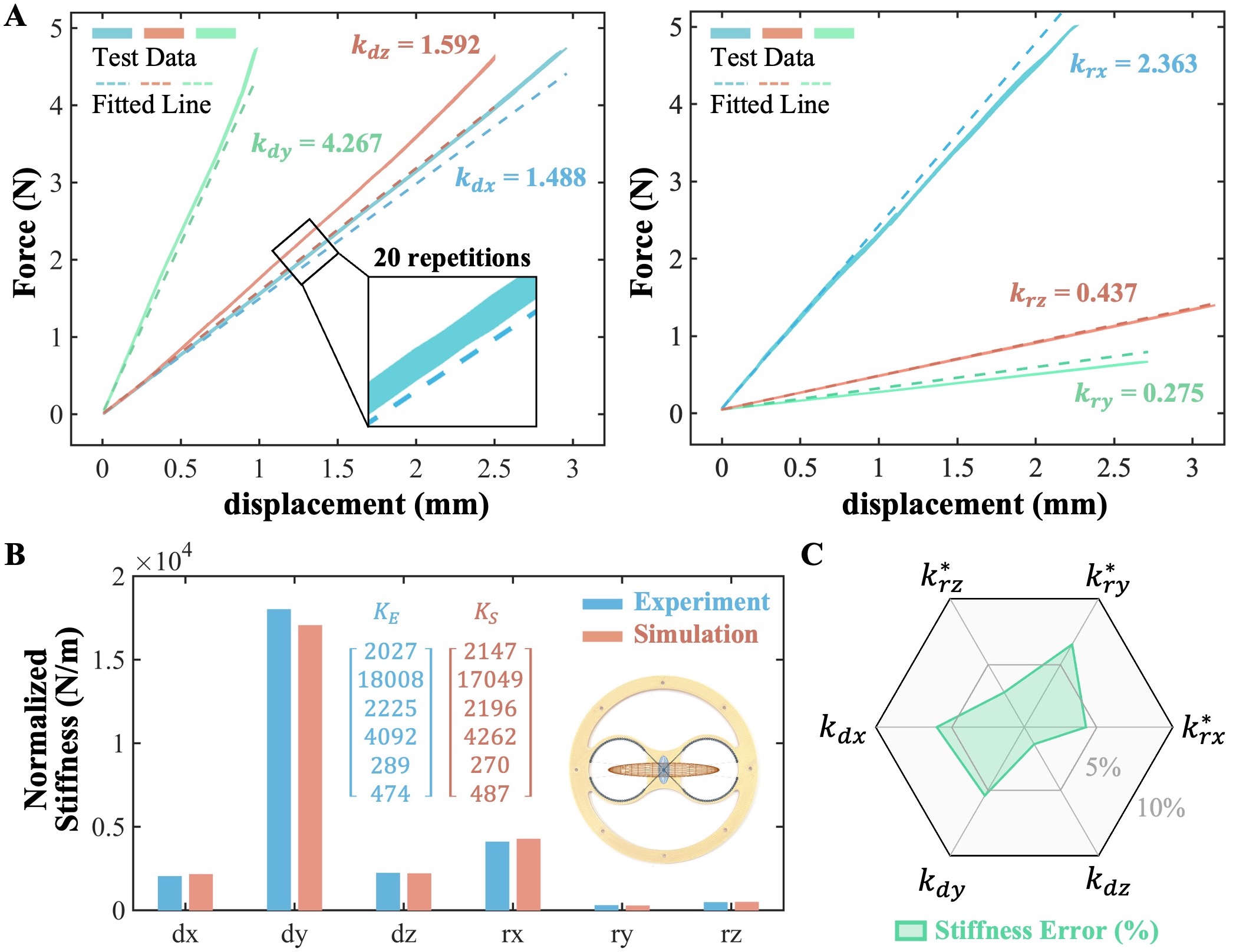}
	\caption{Experimental stiffness results and comparison with the simulated model. 
    \textbf{(A)} Force-displacement curves and linear fitting near the unloaded configuration. 
    \textbf{(B)} Comparison between experimentally identified and simulated stiffness components. 
    \textbf{(C)} Radar plot of relative errors of the six principal stiffness components. }
	\label{fig::stiffRes}
\end{figure}

For each stiffness component, the imposed motion is limited to within $4 \, mm$. 
Each test consists of 20 consecutive repetitions applied at a constant speed of $0.1 \, mm/s$. 
It should be noted that since the load cell adopted for force measurement behaves as a compliant cantilever, the imposed displacement is partly contributed by its own deformation in addition to the compliant module. 
Therefore, its stiffness contribution is explicitly accounted for in the stiffness characterization. 
Specifically, the measured stiffness is the equivalent stiffness of the load cell and the tested module in parallel. 

\subsection{Stiffness Characterization} \label{SubSec::stiff_results}

The measured 20 repetitions of force and displacement data are shown in Fig. \ref{fig::stiffRes}A, depicted by the shaded regions. 
Only small variations are observed among the repeated measurements, with the maximum force spread at the same displacement remaining below $0.1 \, N$ for all six loading configurations. 
This indicates good repeatability of the experimental setup and provides a reliable basis for stiffness characterization. 
Over the prescribed motion range of approximately $4 \, mm$, the force-displacement curves remain close to linear. 
Nevertheless, a moderate change in the local slope can be observed as the deformation increases, with the instantaneous stiffness varying by approximately $20\%$ over the full range. 
To identify the stiffness near the unloaded configuration, the force-displacement data over the initial $0$ to $0.03 \, mm$ range are fitted using linear regression. 
The fitted results are indicated by the dashed lines in Fig. \ref{fig::stiffRes}A. 
For all six components, the linear fits yield $R^{2} \geq 0.99$, demonstrating great linearity. 
To compensate for the compliance of the load cell, its stiffness is independently calibrated by pressing it against a rigid base, resulting in $k_{C} = 5592 \, N/m$. 
The eventual module stiffness $k_{M}$ is then expressed as $k_{M} = \frac{k_{C} \cdot k_{E} }{k_{C} - k_{E}}$ where $k_{E}$ denotes the measured equivalent stiffness, identified from the slope of the force-displacement fitted line. 

The experimentally identified module stiffness is compared with the corresponding simulation results in Fig. \ref{fig::stiffRes}B. 
The simulated results show consistently small deviations from the experimental measurements across all six principal stiffness components, demonstrating the effectiveness of the proposed stiffness model.
The simulated geometry is additionally superimposed on the fabricated module to illustrate their geometric correspondence. 
The relative errors are further depicted by the radar plot in Fig. \ref{fig::stiffRes}C, with all components exhibiting deviations below $6.5 \, \%$ and an average deviation of less than $5 \, \%$. 
It should be noted that the simulated stiffness depends on the elastic properties assigned to the material. 
Since the actual Young's modulus may differ from its nominal value, calibration is adopted for the comparison. 
Therefore, the validated stiffness model provides a reliable basis for implementing task-specific module configurations, constructing compliant interfaces for different tasks and practical applications. 

\section{Demonstrative Experiments}\label{Sec::demos}

\subsection{Flexible Wrist Design}\label{SubSec::wrist_design}

\begin{figure}[!t]
	\centering
	\includegraphics[height=88mm]{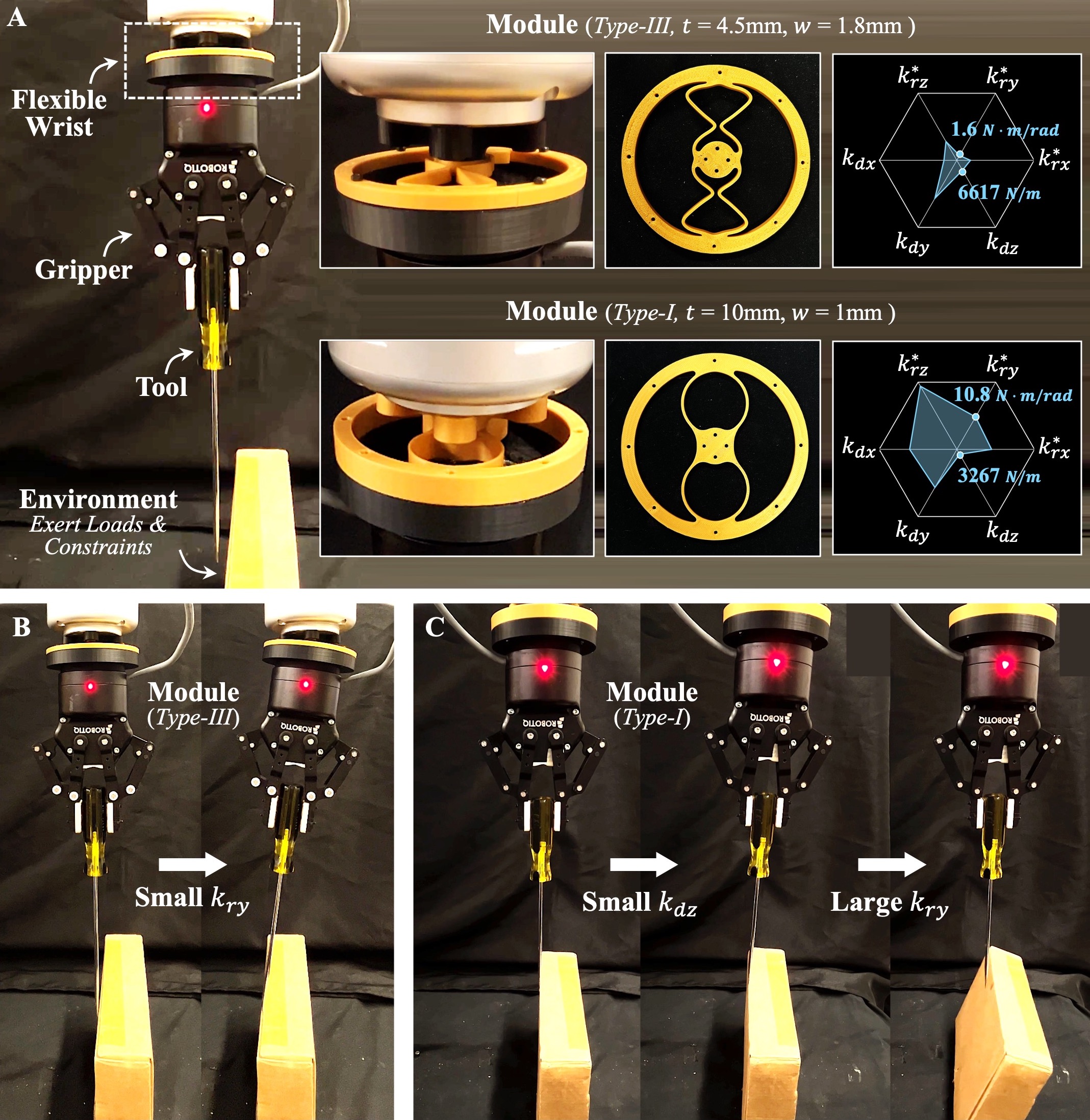}
	\caption{Demonstration of task-dependent contact responses using two compliant wrist configurations. 
    \textbf{(A)} Flexible wrist setup and two compliant module configurations with their corresponding stiffness characteristics. 
    \textbf{(B)} Low rotational stiffness enables the wrist to accommodate the obstacle after contact.  
    \textbf{(C)} Low translational stiffness enables initial compliance after contact, whereas high rotational stiffness eventually tips the obstacle. }
	\label{fig::WristDesign}
\end{figure}

A flexible wrist is developed as an implementation of the proposed stiffness configuration framework. 
As shown in Fig. \ref{fig::WristDesign}A, the compliant module is installed between the robotic arm and the gripper, transmitting interactions between the tool and the environment through the wrist. 
Two module configurations are first implemented to demonstrate the distinct responses induced by different stiffness characteristics, which are depicted by the radar plots for comparison. 
The first employs a \textit{Type-III} module with $\{t, w\} = \{ 4.5 , 1.8\} \, mm$, achieving a low $k_{ry}$ and a low $k_{dz}$, while the second employs a \textit{Type-I} module with $\{t, w\} = \{ 10 , 1\} \, mm$ for a relatively lower $k_{dz}$ and a higher $k_{ry}$. 

The two configurations are subsequently evaluated under the same environmental interaction. 
Specifically, the robotic arm follows the same prescribed motion toward a rigid obstacle, imposing loads and constraints after contact. 
The two wrist responses are illustrated in Fig. \ref{fig::WristDesign}B and C. 
The \textit{Type-III} configuration wrist accommodates the obstacle through angular compliance after contact is established, allowing the grasped tool to continuously reorient as the robotic arm proceeds. 
In contrast, the \textit{Type-I} configuration wrist with a lower $k_{dz}$ initially provides translational compliance after contact, reducing an immediate force increase imposed by the obstacle. 
However, the higher $k_{ry}$ limits further rotational accommodation as the interaction progresses. 
Consequently, the increasing interaction eventually generates sufficient moment to tip the obstacle. 
These results not only validate the capability of realizing distinct passive stiffness characteristics by tuning module parameters, but also demonstrate the significance of task-specific stiffness configurations for achieving desired interaction behaviors. 

\subsection{Wrist Modular Design for Stiffness Configuration}\label{SubSec::task}

\begin{figure}[!t]
	\centering
	\includegraphics[height=62mm]{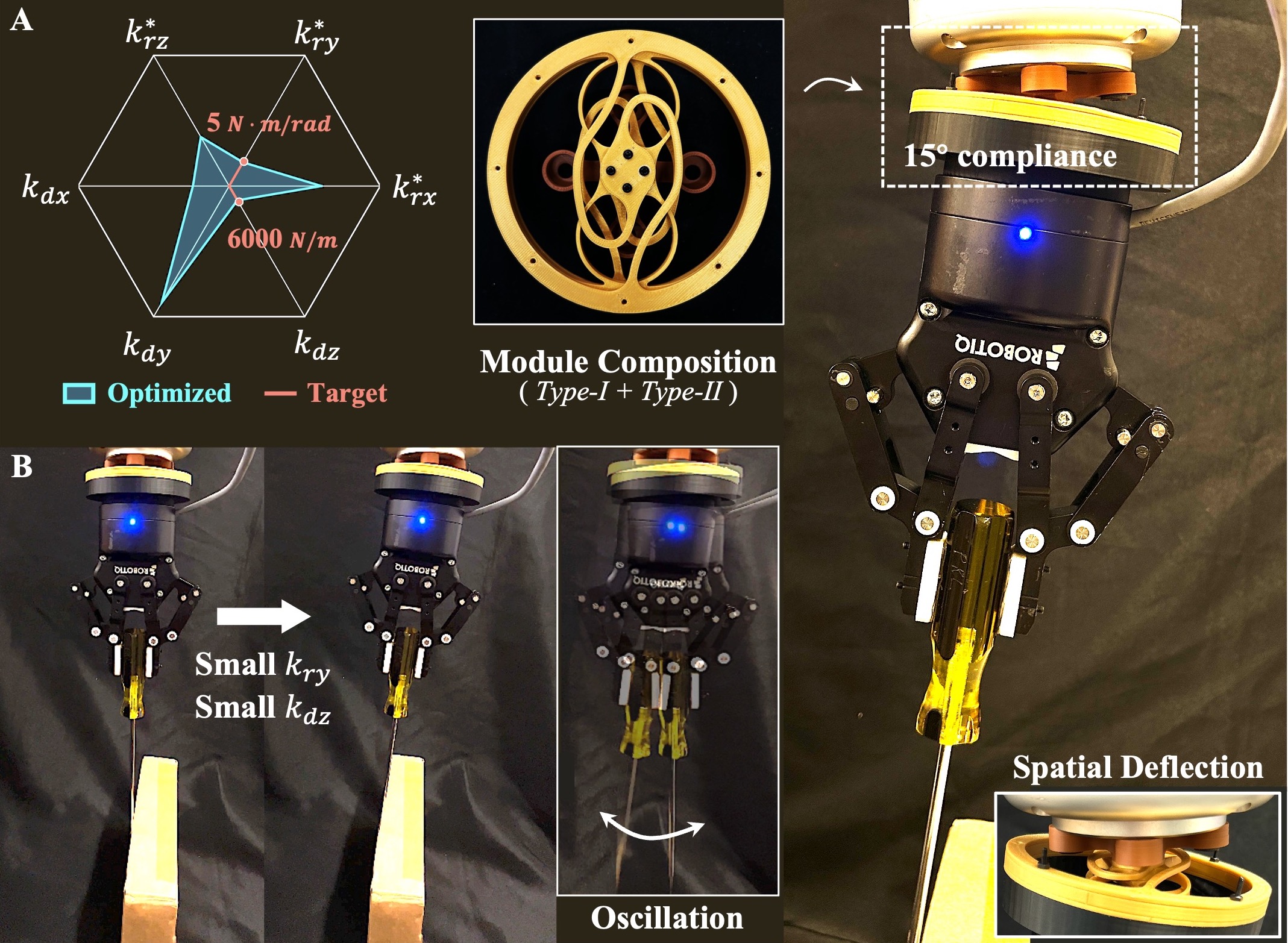}
	\caption{Stiffness configuration and experimental evaluation of the optimized flexible wrist. 
    \textbf{(A)} Optimized stiffness characteristics and corresponding modular composition. 
    \textbf{(B)} Compliance and dynamic performance when interacting with an obstacle at high-speed. }
	\label{fig::wristOpt}
\end{figure}

From the previous experiments, a large rotational stiffness restricts the compliant reorientation of the wrist and consequently tips the contacted object, whereas an excessively low rotational stiffness provides greater accommodation but also introduces noticeable oscillation. 
Besides introducing additional damping, regulating the stiffness is another means of modifying the natural frequency and vibration characteristics of the wrist while retaining the required mechanical compliance. 
This motivates the modular optimization to achieve the stiffness configuration with prescribed values. 

The desired rotational and translational stiffness are specified as $k_{ry} = 5 \, N \cdot m / rad$ and $k_{dz} = 6000 N / m$, which are between the values for the previous two cases. 
Applying the optimization framework introduced in Sec. \ref{SubSec::Opt} yields a solution consisting of a \textit{Type-II} module with $\{t_{1}, w_{1}\} = \{ 1.5 , 2.9\} \, mm$ and a \textit{Type-I} module with $\{t_{2}, w_{2}\} = \{ 4 , 1.4\} \, mm$, both oriented at $0^\circ$. 
It should be noted that for a symmetric arrangement and to prevent interface, two mirrored \textit{Type-II} modules are arranged on the two sides of the \textit{Type-I} module. 
The stiffness radar plot and physical implementation are shown in Fig. \ref{fig::wristOpt}A, demonstrating the optimal solution achieving the prescribed $k_{r_y}$ and $k_{d_z}$ values. 

The optimized wrist is subsequently evaluated under the same obstacle-interaction scenario, as illustrated in \ref{fig::wristOpt}B. 
Similar to the previous low $k_{ry}$ case, angular compliance is observed after the tool contacts the object, allowing an eventually safe pass without tipping the object over. 
However, the optimized wrist exhibits a higher oscillation frequency after passing the object, consistent with the increased prescribed stiffness. 
This indicates that specific stiffness values can be realized through the proposed method, providing a practical means of balancing compliance and dynamic response. 

To further evaluate the compliant response under a severe interaction condition, a maximum vertical offset of $10 \, mm$ is imposed while the robot moves back and forth at $1 \, m/s$. 
The wrist consistently accommodates the obstacle at each contact, showcasing sufficient compliance even under high-speed interaction. 
The experimental demonstrations are provided in the supplementary video. 
A maximum rotation of approximately $15^{\circ}$ is experimentally verified, while further angular compliance remains available beyond the tested condition. 
These results further highlight the value of task-specific stiffness configuration, as different interaction tasks impose distinct stiffness requirements, which can be addressed through the proposed Lego-like modular design and composition approach. 

\section{Discussion and Future Work} \label{Sec::discuss}

\textit{\textbf{Characteristics and Advantages:}}
The key characteristic of the proposed framework is to transform two-dimensional planar geometry optimization into three-dimensional modular composition with parameter optimization. 
The approach results in rapid inverse design and low-cost manufacturing due to its planar nature and predetermined geometry. 
It also gives a larger design region for one module. 
The stacked architecture remains compact. 
Moreover, the framework is readily extensible since modules with new geometries can be incorporated into the set, where the same modeling and optimization strategy can be generally applied. 

\textit{\textbf{Limitations:}}
The present stiffness configuration framework is developed under two main assumptions: the stiffness is evaluated under the unloaded case, and the target stiffness matrix is diagonal. 
In addition, although multiple planar modules enlarge the attainable stiffness design region, the resulting stiffness characteristics are still fundamentally constrained by the mechanics of the underlying planar compliant structures. 
The six principal stiffness components are partially coupled such that an arbitrary configuration cannot generally be achieved even with an expanded module set. 
When the desired stiffness lies outside the attainable design region, the optimization can only provide a compromise according to the given weights. 
Stacking also introduces additional mechanical effects. 
As the stack thickness increases, the offset between individual modules and the reference rotation center causes translational stiffness to contribute to the equivalent rotational stiffness, thereby modifying the original characteristics associated with its own center. 
Moreover, different module geometries may undergo different out-of-plane deformations during rotation, which can lead to geometric interference and alter the effective stiffness. 

\textit{\textbf{Future Extension:}}
Firstly, the module set can be expanded to push the boundary of the stiffness design space. 
Additional planar geometries with complementary stiffness characteristics can be introduced, while spatial compliant chains may further extend the configuration capability beyond the limitations of planar mechanisms. 
Secondly, by introducing asymmetric compliant chains and module layouts, future designs could incorporate controllable coupling terms and enable configuration of a non-diagonal stiffness matrix. 
Such capability may satisfy more diverse stiffness requirements, including the realization of remote-center-compliance (RCC) mechanisms with passive compliance. 
Furthermore, the broad design space offers the possibility of tailoring the stiffness variation during deformation, which would enable task-specific regulation of nonlinear stiffness characteristics for more complex physical interactions. 

\section{Conclusion}\label{Sec::conclusion}

This paper presented a Lego-like stiffness configuration approach using stackable planar compliant modules. 
Parallel composition of parameterized modules enables task-specific stiffness regulation without redesigning the complete mechanism. 
A unified stiffness model and a two-stage optimization method were developed for quantitative analysis and rapid inverse design. 
Experiments validated the simulated results, and a flexible wrist with optimized prescribed stiffness demonstrated desired compliance and dynamic performance during high-speed interaction. 
Overall, the proposed framework transforms in-plane geometry optimization into three-dimensional stacking and parameter optimization of predefined planar compliant modules, providing a systematic basis for constructing a broader range of flexible interfaces, beyond the wrist, to tailor the stiffness characteristics according to task-specific interaction requirements. 

\bibliographystyle{IEEEtran}
\normalem
\bibliography{Ref}

\vfill

\end{sloppypar}
\end{document}